\documentclass[sigconf, nonacm]{acmart}
\usepackage{epsfig}
\usepackage{graphicx}
\usepackage{amsmath}
\usepackage{amssymb}

\usepackage{subfigure}
\usepackage{bm}
\usepackage[utf8]{inputenc}
\usepackage[english]{babel}
\usepackage{comment}
\usepackage{color}
\usepackage{etoolbox}
\usepackage{booktabs}
\usepackage{multirow}
\usepackage{dblfloatfix}

\usepackage{algorithm}
\usepackage{algorithmic}
\usepackage[algo2e,ruled,linesnumbered]{algorithm2e}
\usepackage{amsthm}
\usepackage{makecell}

\usepackage{wrapfig}
\usepackage{caption}
\usepackage{wasysym}
\usepackage{multicol}

\newbool{inccomment}
\setbool{inccomment}{true}
\newcommand{\hl}[1]{\ifbool{inccomment}{{\color{magenta}#1}}{}}
\newcommand{\ww}[1]{\ifbool{inccomment}{{\color{blue} #1}}{}}
\newcommand{\yc}[1]{\ifbool{inccomment}{{\color{red} #1}}{}}

\newcommand{\autogrow}{\textit{AutoGrow}}
\newcommand{\resnet}{\textit{ResNet}}
\newcommand{\resnets}{\textit{ResNets}}
\newcommand{\ie}{\textit{i.e.}}
\newcommand{\eg}{\textit{e.g.}}

\usepackage{hyperref}
\usepackage{url}

\AtBeginDocument{%
  \providecommand\BibTeX{{%
    \normalfont B\kern-0.5em{\scshape i\kern-0.25em b}\kern-0.8em\TeX}}}

\setcopyright{acmcopyright}
\copyrightyear{2020}
\acmYear{2020}
\acmDOI{10.1145/xxxxxxx.xxxxxxx}

\acmConference[KDD '20]{KDD '20: Knowledge Discovery and Data Mining}{August 22--27, 2020}{San Diego, CA}
\acmBooktitle{KDD '20: Knowledge Discovery and Data Mining, August 22--27, 2020, San Diego, CA}
\acmPrice{15.00}
\acmISBN{978-1-4503-XXXX-X/18/06}

\acmSubmissionID{rfp0820}

\begin{document}
\fancyhead{}

\title{\autogrow{}: Automatic Layer Growing \\ in Deep Convolutional Networks}

\author{Wei Wen}
\email{wei.wen@alumni.duke.edu}
\affiliation{%
	\institution{Duke University}
}

\author{Feng Yan}
\email{fyan@unr.edu}
\affiliation{%
  \institution{University of Nevada, Reno}
}

\author{Yiran Chen}
\email{yiran.chen@duke.edu}
\affiliation{%
	\institution{Duke University}
}

\author{Hai Li}
\email{hai.li@duke.edu}
\affiliation{%
	\institution{Duke University}
}

\renewcommand{\shortauthors}{Wen, et al.}

\begin{abstract}
  Depth is a key component of Deep Neural Networks (DNNs), however, designing depth is heuristic and requires many human efforts. 
  We propose \autogrow{} to automate depth discovery in DNNs: 
  starting from a shallow seed architecture, \autogrow{} grows new layers if the growth improves the accuracy; otherwise, stops growing and thus discovers the depth. 
  We propose robust growing and stopping policies to generalize to different network architectures and datasets.
  Our experiments show that by applying the same policy to different network architectures, \autogrow{} can always discover near-optimal depth on various datasets of MNIST, FashionMNIST, SVHN, CIFAR10, CIFAR100 and ImageNet. 
  For example, in terms of accuracy-computation trade-off, \autogrow{} discovers a better depth combination in \resnets{} than human experts.
  Our \autogrow{} is efficient. It discovers depth within similar time of training a single DNN.
  Our code is available at \url{https://github.com/wenwei202/autogrow}.
\end{abstract}

%

\keywords{automated machine learning; neural architecture search; depth; growing; neural networks; automation}


\maketitle

\section{Introduction}
\label{sec:intro}
Layer depth is one of the decisive factors of the success of Deep Neural Networks (DNNs).
For example, image classification accuracy keeps improving as the depth of network models grows~\citep{krizhevsky2012imagenet,simonyan2014very,szegedy2015going,he2016deep,huang2017densely}.
Although shallow networks cannot ensure high accuracy, DNNs composed of too many layers may suffer from over-fitting and convergence difficulty in training.
How to obtain the optimal depth for a DNN still remains mysterious.
For instance, \resnet{}-152~\citep{he2016deep} uses $3$, $8$, $36$ and $3$ residual blocks under output sizes of $56 \times 56$, $28 \times 28$, $14 \times 14$ and $7 \times 7$, respectively, which don't show an obvious quantitative relation. 
In practice, people usually reply on some heuristic trials and tests to obtain the depth of a network: 
they first design a DNN with a specific depth and then train and evaluate the network on a given dataset; finally, they change the depth and repeat the procedure until the accuracy meets the requirement.  
Besides the high computational cost induced by the iteration process, such trial \& test iterations must be repeated whenever dataset changes.
In this paper, we propose \autogrow{} that can automate depth discovery given a layer architecture.
We will show that \autogrow{} generalizes to different datasets and layer architectures.


There are some previous works which add or morph layers to increase the depth in DNNs.
VggNet~\citep{simonyan2014very} and DropIn~\citep{smith2016gradual} added new layers into shallower DNNs;
Network Morphism~\citep{wei2016network, wei2017modularized, chen2015net2net} morphed each layer to multiple layers to increase the depth meanwhile preserving the function of the shallower net.
Table~\ref{tab:work_comparison} summarizes differences in this work. Their goal was to overcome difficulty of training deeper DNNs or accelerate it. Our goal is to automatically find an optimal depth.
Moreover, previous works applied layer growth by once or a few times at pre-defined locations to grow a pre-defined number of layers; in contrast, ours automatically learns the number of new layers and growth locations without limiting growing times.
We will summarize more related works in Section~\ref{sec:related}.

\begin{figure}[b]
	\centering
	\includegraphics[width=1.0\columnwidth]{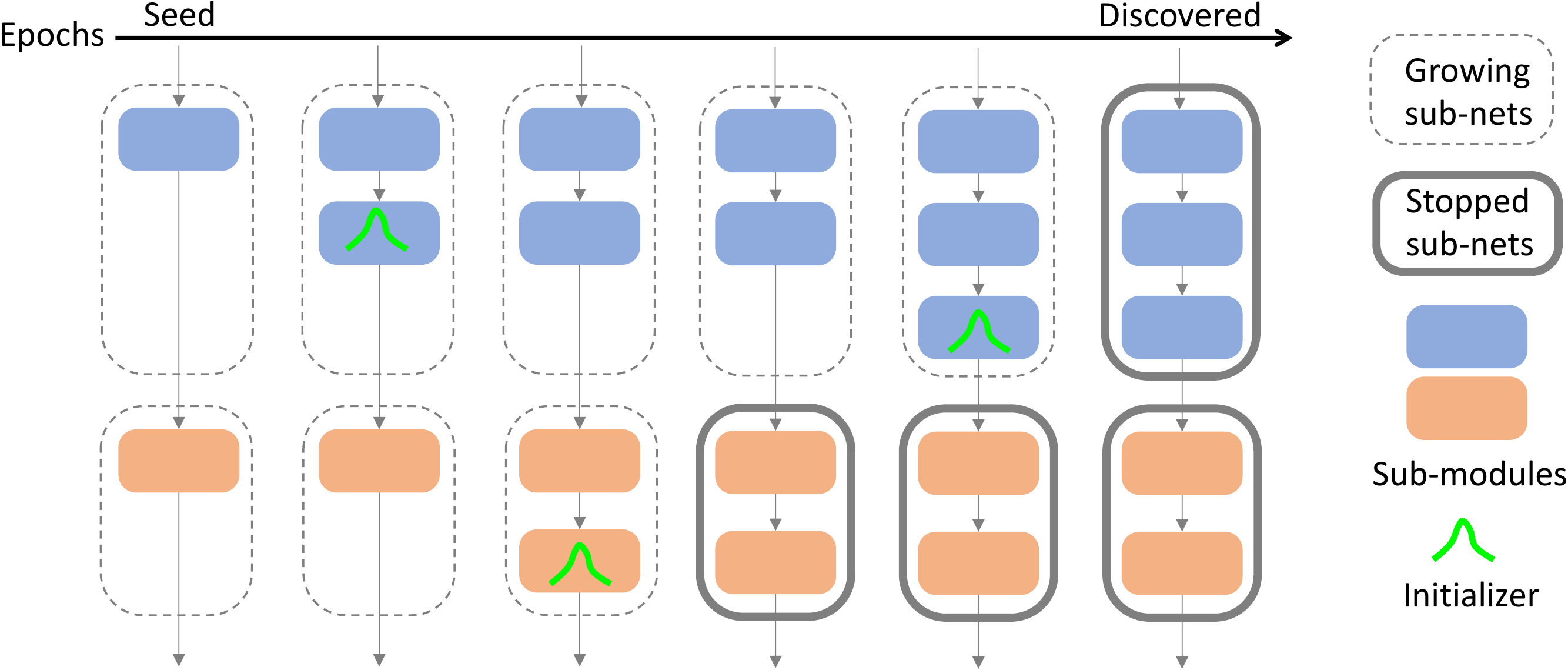}
	\captionof{figure}{A simple example of \autogrow{}.}
	\label{fig:autogrow}
\end{figure}

\begin{table}[b]
	\centering
	\captionof{table}{Comparison with previous works about layer growth.}
	\label{tab:work_comparison}
	\resizebox{.8\columnwidth}{!}{
		\begin{tabular}{rrr}
			\toprule
			& Previous works &  Ours \\
			\midrule 
			Goal & Ease training & Depth automation \\
			Times & Once or a few & Unlimited \\
			Locations & Human defined & Learned \\
			Layer \# & Human defined & Learned \\
			\bottomrule
		\end{tabular}
	}
\end{table}

Figure~\ref{fig:autogrow} illustrates an example of \autogrow{}. 
It starts from the shallowest backbone network and gradually grows \textit{sub-modules} (A \textit{sub-module} can be one or more layers, \eg{}, a residual block); 
the growth stops once a stopping policy is satisfied.
We studied multiple initializers of new layers and multiple growing policies, and surprisingly find that:
(1) a random initializer works equally or better than complicated Network Morphism;
(2) it is more effective to grow \textit{before} a shallow net converges.
We hypothesize that this is because a converged shallow net is an inadequate initialization for training deeper net, while random initialization can help to escape from a bad starting point.
Motivated by this, we intentionally avoid full convergence during the growing by using (1) random initialization of new layers, (2) a constant large learning rate, and (3) a short growing interval.
Our contributions are:
\begin{itemize}
\item We propose \autogrow{} to automate DNN layer growing and depth discovery. \autogrow{} is very robust. 
With the same hyper-parameters, it adapts network depth to various datasets including MNIST, FashionMNIST, SVHN, CIFAR10, CIFAR100 and ImageNet. 
Moreover, \autogrow{} can also discover shallower DNNs when the dataset is a subset. 
\item \autogrow{} demonstrates high efficiency and scales up to ImageNet, because the layer growing is as fast as training a single DNN. 
On ImageNet, it discovers a new \resnets{} with better trade-off between accuracy and computation complexity.
\item We challenge the idea of Network Morphism, as random initialization works equally or better when growing layers.
\item We find that it is beneficial to rapidly grow layers before a shallower net converge, contradicting previous intuition.
\end{itemize}

\section{\autogrow{} -- A Depth Growing Algorithm}
\label{sec:method}


\begin{algorithm}
	\SetArgSty{textnormal}
	\caption{\autogrow{} Algorithm.} 
	\label{algo:autogrow}
	\begin{algorithmic}
	\STATE \textbf{Input:}
	
	\STATE A seed shallow network $g(\mathcal{X}_0)$ composed of $M$ sub-networks $\mathbb{F} =\left\{  \bm{f}_i \left( \cdot ;\mathbb{W}_i \right) : i = 0 \dots M-1 \right\} $, where each sub-network has only one sub-module (a dimension reduction sub-module);
	
	\STATE an epoch interval $K$ to check growing and stopping policies;
	
	\STATE the number of fine-tuning epochs $N$ after growing.
	
	\STATE
	\STATE \textbf{Initialization:} 

	\STATE A Circular Linked List of sub-networks under growing: \\
	$\texttt{subNetList} = \underleftarrow{\bm{f}_0 \left( \cdot ;\mathbb{W}_0 \right) \rightarrow  \cdots \rightarrow \bm{f}_{M-1} \left( \cdot ;\mathbb{W}_{M-1} \right) }$;

	\STATE The current growing sub-network: \\
	$\texttt{growingSub} = \texttt{subNetList.head()} = \bm{f}_0 \left( \cdot ;\mathbb{W}_0 \right)$;
	
	\STATE The last grown sub-network: \texttt{grownSub} = \texttt{None};
	
	\STATE
	\STATE \textbf{Process:} 
	
	\STATE \textit{\# if there exist growing sub-network(s)}
	\WHILE{\texttt{subNetList.size()>0}}
	\STATE 
		
		\texttt{train($g(\mathcal{X}_0), K$)} \textit{\# train the whole network $g(\mathcal{X}_0)$ for $K$ epochs}
		
		\IF{\texttt{meetStoppingPolicy()}}
			\STATE \textit{\# remove a sub-network from the growing list}
			\STATE \texttt{subNetList.delete(grownSub)};
			
		\ENDIF

		\IF{\texttt{meetGrowingPolicy() and} \texttt{subNetList.size()>0}}
			\STATE
			\textit{\# the current growing sub-network $\texttt{growingSub} == \bm{f}_i \left( \cdot ;\mathbb{W}_i \right) $ }

			$\mathbb{W}_i = \mathbb{W}_i \cup \mathcal{W}$ \textit{\# stack a sub-module on top of} $\bm{f}_i \left( \cdot ;\mathbb{W}_i \right)$

			\texttt{initializer($\mathcal{W}$)}; \textit{\# initialize the new sub-module} $\mathcal{W}$
			
			\texttt{grownSub} $=$ \texttt{growingSub};
			
			\texttt{growingSub} $=$  \texttt{subNetList.next(growingSub)}; 
			
		\ENDIF
	\ENDWHILE
	
	\STATE Fine-tune the discovered network $g(\mathcal{X}_0)$ for $N$ epochs;
	
	\STATE
	\STATE \textbf{Output:}

	\STATE A trained neural network $g(\mathcal{X}_0)$ with learned depth.
	\end{algorithmic}
\end{algorithm}

Figure~\ref{fig:autogrow} gives an overview of the proposed \autogrow{}. 
In this paper, we use \textit{network}, \textit{sub-networks}, \textit{sub-modules} and \textit{layers} to describe the architecture hierarchy. 
A \textit{network} is composed of a cascade of \textit{sub-networks}.
A \textit{sub-network} is composed of \textit{sub-modules}, which typical share the same output size. 
A \textit{sub-module} (\textit{e.g.} a residual block) is an elementary growing block composed of one or a few \textit{layers}.
In this section, we rigorously formulate a generic version of \autogrow{} which will be materialized in subsections.
A deep convolutional \textit{network} $g(\mathcal{X}_0)$ is a cascade of \textit{sub-networks} by composing functions as
$g(\mathcal{X}_0) = l \left(  \bm{f}_{M-1}  \left( \bm{f}_{M-2}  \left( \cdots \bm{f}_1  \left(  \bm{f}_0 \left( \mathcal{X}_0 \right) \right) \cdots \right) \right) \right)$,
where $\mathcal{X}_0$ is an input image, 
$M$ is the number of sub-networks, 
$l(\cdot)$ is a loss function, 
and $\mathcal{X}_{i+1} = \bm{f}_i \left( \mathcal{X}_i \right)$ is a \textit{sub-network} that operates on an input image or a feature tensor $\mathcal{X}_i \in \mathbb{R}^{c_i \times h_i \times w_i} $.
Here, $c_i$ is the number of channels, and $h_i$ and $w_i$ are spatial dimensions.
$\bm{f}_i \left( \mathcal{X}_i \right)$ is a 
simplified notation
of $\bm{f}_i \left( \mathcal{X}_i ;\mathbb{W}_i \right)$, where $\mathbb{W}_i$ is a set of \textit{sub-modules}' parameters within the $i$-th \textit{sub-network}. 
Thus $\mathbb{W} = \left\{ \mathbb{W}_i : i=0 \dots M-1 \right\}$ denotes the whole set of parameters in the DNN.
To facilitate growing, the following properties are supported within a sub-network: (1) the first sub-module usually reduces the size of input feature maps, \textit{e.g.}, using pooling or convolution with a stride; and
(2) all sub-modules in a sub-network maintain the same output size. 
As such, our framework can support popular networks,
including \textit{VggNet}-like plain networks~\citep{simonyan2014very}, \textit{GoogLeNet}~\citep{szegedy2015going}, \resnets{}~\citep{he2016deep} and \textit{DenseNets}~\citep{huang2017densely}. 
In this paper, we select \resnets{} and \textit{VggNet}-like nets as representatives of DNNs with and without shortcuts, respectively.

With above notations, Algorithm~\ref{algo:autogrow} rigorously describes the \autogrow{} algorithm.
In brief, \autogrow{} starts with the shallowest net where every sub-network has only one sub-module for spatial dimension reduction.
\autogrow{} loops over all growing sub-networks in order.
For each sub-network, \autogrow{} stacks a new sub-module. When the new sub-module does not improve the accuracy, the growth in corresponding sub-network will be permanently stopped.
The details of our method will be materialized in the following subsections.

\subsection{Seed Shallow Networks and Sub-modules}
In this paper, in all datasets except ImageNet, we explore growing depth for four types of DNNs: 
\begin{itemize}
	\item \texttt{Basic3ResNet}: the same \resnet{} used for CIFAR10 in~\cite{he2016deep}, which has $3$ residual \textit{sub-networks} with output spatial sizes of $32 \times 32$, $16 \times 16$ and $8 \times 8$, respectively;
	\item \texttt{Basic4ResNet}: a variant of \resnet{} used for ImageNet in~\cite{he2016deep} built by basic residual blocks (each of which contains two convolutions and one shortcut).
	There are $4$ \textit{sub-networks} with output spatial sizes of $32 \times 32$, $16 \times 16$, $8 \times 8$ and $4 \times 4$, respectively;
	\item \texttt{Plain3Net}: a \textit{VggNet}-like plain net by removing shortcuts in \texttt{Basic3ResNet};
	\item \texttt{Plain4Net}: a \textit{VggNet}-like plain net by removing shortcuts in \texttt{Basic4ResNet}.
\end{itemize}


In \autogrow{}, the architectures of seed shallow networks and sub-modules are pre-defined.
In plain DNNs, a \textit{sub-module} is a stack of convolution, Batch Normalization and ReLU;
in residual DNNs, a \textit{sub-module} is a residual block.
In \autogrow{}, a \textit{sub-network} is a stack of all sub-modules with the same output spatial size. 
Unlike~\cite{he2016deep} which manually designed the depth, \autogrow{} starts from a seed architecture in which each sub-network has only one sub-module and automatically learns the number of sub-modules.

On ImageNet, we apply the same backbones in~\cite{he2016deep} as the seed architectures. A seed architecture has only one sub-module under each output spatial size.
For a \resnet{} using basic residual blocks or bottleneck residual blocks~\citep{he2016deep}, we respectively name it as \texttt{Basic4ResNet} or \texttt{Bottleneck4ResNet}. 
\texttt{Plain4Net} is also obtained by removing shortcuts in \texttt{Basic4ResNet}. 

\subsection{Sub-module Initializers}
Here we explain how to initialize a new sub-module $\mathcal{W}$ mentioned in Algorithm~\ref{algo:autogrow} ($\texttt{initializer}(\mathcal{W})$).
Network Morphism changes DNN architecture meanwhile preserving the loss function via special initialization of new layers, that is,
\begin{equation}
g(\mathcal{X}_0;\mathbb{W}) = g(\mathcal{X}_0;\mathbb{W} \cup \mathcal{W})~\forall \mathcal{X}_0.
\end{equation}
A residual sub-module shows a nice property: when stacking a residual block and initializing the last Batch Normalization layer as zeros, the function of the shallower net is preserved but the DNN is morphed to a deeper net. Thus, Network Morphism can be easily implemented by this zero initialization (\texttt{ZeroInit}). 

In this work, all layers in $\mathcal{W}$ are initialized using default randomization, except for a special treatment of the last Batch Normalization layer in a \textit{residual} sub-module.
Besides \texttt{ZeroInit}, we propose a new \texttt{AdamInit} for Network Morphism. 
In \texttt{AdamInit}, we freeze all parameters except the last Batch Normalization layer in $\mathcal{W}$, and then use Adam optimizer~\citep{kingma2014adam} to optimize the last Bath Normalization for maximum $10$ epochs till the training accuracy of the deeper net is as good as the shallower one. 
After \texttt{AdamInit}, all parameters are jointly optimized. 
We view \texttt{AdamInit} as a Network Morphism because the training loss is similar after \texttt{AdamInit}. 
We empirically find that \texttt{AdamInit} can usually find a solution in less than $3$ epochs.
We also study random initialization of the last Batch Normalization layer using uniform (\texttt{UniInit}) or Gaussian (\texttt{GauInit}) noises with a standard deviation $1.0$.
We will show that \texttt{GauInit} obtains the best result, challenging the idea of Network Morphism~\citep{wei2016network, wei2017modularized, chen2015net2net}.

\subsection{Growing and Stopping Policies}
In Algorithm~\ref{algo:autogrow}, a growing policy refers to $\texttt{meetGrowingPolicy()}$, which returns true when the network should grow a sub-module.
Two growing policies are studied here:
\begin{enumerate}
\item Convergent Growth:  \texttt{meetGrowingPolicy()} returns true when the improvement of validation accuracy is less than $\tau$ in the last $K$ epochs. 
That is, in Convergent Growth, \autogrow{} only grows when current network has converged. 
This is a similar growing criterion adopted in previous works~\citep{elsken2017simple,cai2018efficient,cai2018path}.
\item Periodic Growth:  \texttt{meetGrowingPolicy()} always returns true, that is, the network always grows every $K$ epochs. Therefore, $K$ is also the growing period.
In the best practice of \autogrow{}, $K$ is small (\eg{} $K=3$) such that it grows before current network converges.
\end{enumerate}
Our experiments will show that Periodic Growth outperforms Convergent Growth. 
We hypothesize that a fully converged shallower net is an inadequate initialization to train a deeper net.
We will perform experiments to test this hypothesis and visualize optimization trajectory to illustrate it.

A stopping policy denotes $\texttt{meetStoppingPolicy()}$ in Algorithm \ref{algo:autogrow}.
When Convergent Growth is adopted, \texttt{meetStoppingPolicy()} returns true if a recent growth does not improve validation accuracy more than $\tau$ within $K$ epochs.
We use a similar stopping policy for Periodic Growth; however, as it can grow rapidly with a small period $K$ (\eg{} $K=3$) before it converges, we use a larger window size $J$ for stop. Specifically, when Periodic Growth is adopted, \texttt{meetStoppingPolicy()} returns true when the validation accuracy improves less than $\tau$ in the last $J$ epochs, where $J \gg K$. 


Hyper-parameters $\tau$, $J$ and $K$ control the operation of \autogrow{} and can be easily setup and generalize well.
$\tau$ denotes the significance of accuracy improvement for classification.
We simply set $\tau=0.05\%$ in all experiments. 
$J$ represents how many epochs to wait for an accuracy improvement before stopping the growth of a sub-network.
It is more meaningful to consider stopping when the new net is trained to some extent. 
As such, we set $J$ to the number of epochs $T$ under the largest learning rate when training a baseline.
$K$ means how frequently \autogrow{} checks the polices. 
In Convergent Growth, we simply set $K=T$, which is long enough to ensure convergence.
In Periodic Growth, $K$ is set to a small fraction of $T$ to enable fast growth before convergence;
more importantly, $K=3$ is very robust to all networks and datasets.
Therefore, all those hyper-parameters are very robust and strongly correlated to design considerations.

\section{Experiments}

\begin{table}[b]
  \caption{ Network Morphism tested on CIFAR10. }
  \label{tab:morphism-one2one}
  \centering
  \resizebox{1.\columnwidth}{!}{
  	\begin{tabular}{cccccc}
  		\toprule
  		 net backbone & shallower & deeper &  initializer & accu \% & $\Delta$\textsuperscript{$*$}  \\
  		
  		\midrule 
  		
  			\multirow{2}{*}{\texttt{Basic3ResNet}} & \multirow{2}{*}{\texttt{3-3-3}}  & \multirow{2}{*}{\texttt{5-5-5}} & \texttt{ZeroInit} & 92.71 & -0.77 \\
  		& & & \texttt{AdamInit} &  92.82  & -0.66 \\
  		
  		\midrule 
  		
  		\multirow{2}{*}{\texttt{Basic3ResNet}} & \multirow{2}{*}{\texttt{5-5-5}}  & \multirow{2}{*}{\texttt{9-9-9}} & \texttt{ZeroInit} & 93.64 & -0.27  \\
  		 & & & \texttt{AdamInit} &  93.53 & -0.38 \\
  		
  		\midrule 
  		\multirow{2}{*}{\texttt{Basic4ResNet}} & \multirow{2}{*}{\texttt{1-1-1-1}}  & \multirow{2}{*}{\texttt{2-2-2-2}} & \texttt{ZeroInit} & 94.96 & -0.37  \\
  		& & & \texttt{AdamInit} &  95.17 & -0.16 \\
  		
  		\bottomrule
  		\multicolumn{6}{l}{\textsuperscript{$*$} $\Delta = $ (accuracy of Network Morphism) $-$ (accuracy of training from scratch) }\\
  	\end{tabular}
  }
\end{table} 
\begin{table*}[t]
  \caption{ Ablation study of \textit{c-AutoGrow}. }
  \label{tab:autogrow-morphism}
  \centering
  \resizebox{2.0\columnwidth}{!}{
  	\begin{tabular}{cccccr}
  		\toprule
  		dataset & \begin{tabular}{@{}c@{}}learning \\ rate\end{tabular} &  initializer & found net\textsuperscript{$\dagger$} & accu \%  & $\Delta$\textsuperscript{$*$} \\
  		\midrule 
  		\multirow{6}{*}{CIFAR10} & \underline{\textit{staircase}}  & \texttt{ZeroInit} & \texttt{2-3-6} & 91.77 & -1.06 \\
  		& \underline{\textit{staircase}}  & \texttt{AdamInit} & \texttt{3-4-3} & 92.21 & -0.59 \\
  		
  		\cmidrule{2-6}
  		& constant  & \texttt{ZeroInit} & \texttt{2-2-4} & 92.23  & 0.16\\
  		& constant  & \texttt{AdamInit} & \texttt{3-4-4} &  92.60 & -0.41 \\ 
  		& constant  & \texttt{UniInit} & \texttt{3-4-4} & 92.93  & -0.08\\
  		& \textbf{constant}  & \textbf{\texttt{GauInit}} & \textbf{\texttt{2-4-3}} & \textbf{93.12}  & \textbf{0.55} \\

  		\bottomrule
  		\multicolumn{6}{l}{\textsuperscript{$\dagger$} \texttt{Basic3ResNet} }\\
  	\end{tabular}
  \quad
  \begin{tabular}{cccccr}
  	\toprule
  	dataset & \begin{tabular}{@{}c@{}}learning \\ rate\end{tabular} &  initializer & found net\textsuperscript{$\dagger$} & accu \%  & $\Delta$\textsuperscript{$*$} \\
  	
  	\midrule 
  	
  	\multirow{6}{*}{CIFAR100} & \underline{\textit{staircase}}  & \texttt{ZeroInit} &  \texttt{4-3-4} & 70.04  & -0.65 \\
  	& \underline{\textit{staircase}}  & \texttt{AdamInit} &  \texttt{3-3-3} & 69.85  & -0.65 \\
  	
  	\cmidrule{2-6}
  	
  	& constant  & \texttt{ZeroInit} & \texttt{3-2-4} & 70.22   &  0.35\\
  	& constant  & \texttt{AdamInit} & \texttt{3-3-3} & 70.00 & -0.50 \\
  	& constant  & \texttt{UniInit} & \texttt{4-4-3} & 70.39   & 0.36 \\
  	& \textbf{constant}  & \textbf{\texttt{GauInit}} & \textbf{\texttt{3-4-3}} & \textbf{70.66} & \textbf{0.91}\\
  	
  	\bottomrule
  	\multicolumn{6}{l}{\textsuperscript{$*$} $\Delta = $ (accuracy of \textit{c-AutoGrow}) $-$ (accuracy of training from scratch) }
  \end{tabular}
  
  }
\end{table*} 

In this paper, we use \texttt{Basic3ResNet-2-3-2}, for instance, to denote a model architecture which contains $2$, $3$ and $2$ sub-modules in the first, second and third sub-networks, respectively. 
Sometimes we simplify it as \texttt{2-3-2} for convenience.
\autogrow{} always starts from the shallowest depth of \texttt{1-1-1} and uses the maximum validation accuracy as the metric to guide growing and stopping.
All DNN baselines are trained by SGD with momentum $0.9$ using staircase learning rate.
The initial learning rate is $0.1$ in \resnets{} and $0.01$ in plain networks.
On ImageNet, baselines are trained using  batch size $256$ for $90$ epochs, within which learning rate is decayed by $0.1 \times$ at epoch $30$ and $60$.
In all other smaller datasets, baselines are trained using  batch size $128$ for $200$ epochs and learning rate is decayed by $0.1 \times$ at epoch $100$ and $150$.

Our early experiments followed prior wisdom by growing layers with Network Morphism~\citep{wei2016network, wei2017modularized, chen2015net2net, elsken2017simple,cai2018efficient,cai2018path}, \ie{}, \autogrow{} with \texttt{ZeroInit} (or \texttt{AdamInit}) and Convergent Growth policy; however, it stopped early with very shallow DNNs, failing to find optimal depth.
We hypothesize that a converged shallow net with Network Morphism gives a bad initialization to train a deeper neural network.
Section~\ref{exp:morphism} experimentally test that the hypothesis is valid. 
To tackle this issue, we intentionally avoid convergence during growing by three simple solutions, which are evaluated in Section~\ref{exp:autogrow}.
Finally, Section~\ref{exp:adaptation} and Section~\ref{exp:imagenet} include extensive experiments to show the effectiveness of our final \autogrow{}.


\subsection{Suboptimum of Network Morphism and Convergent Growth}
\label{exp:morphism}

\begin{figure}
	\centering
	\includegraphics[width=1.\columnwidth]{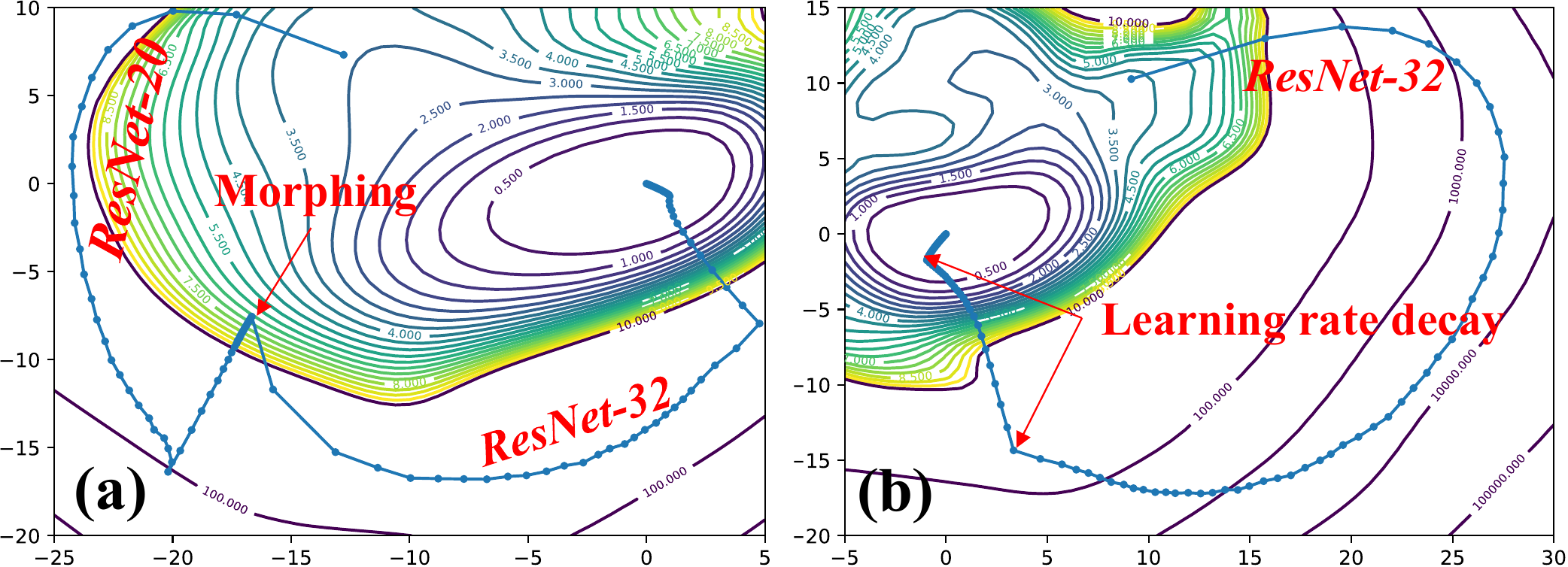}
	\caption{An optimization trajectory comparison between (a) Network Morphism and (b) training from scratch.}
	\label{fig:trajectory-morphism}
\end{figure}

\begin{figure*}
	\centering
	\includegraphics[width=2.0\columnwidth]{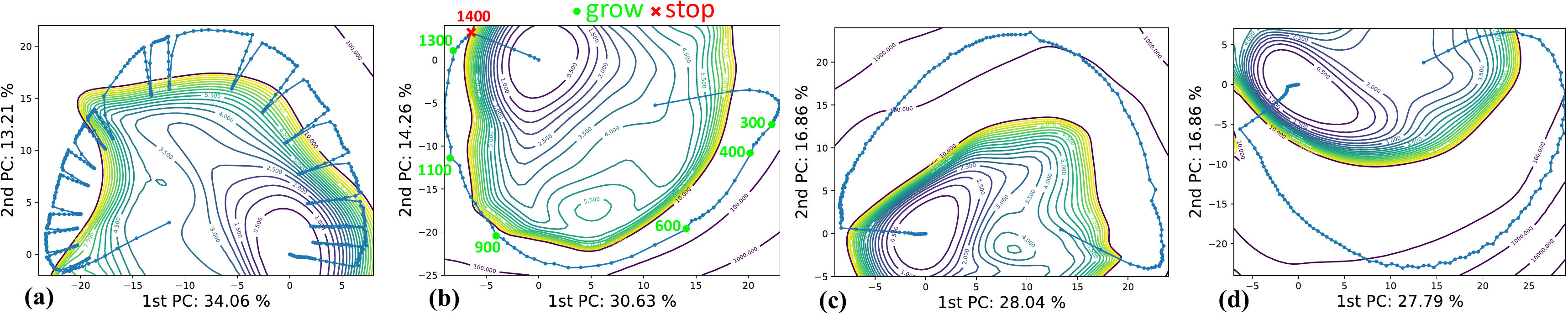}
	\caption{Optimization trajectory of \autogrow{}, tested by \texttt{Basic3ResNet} on CIFAR10.  
		(a) \textit{c-AutoGrow} with staircase learning rate and \texttt{ZeroInit} during growing; 
		(b) \textit{c-AutoGrow} with constant learning rate and \texttt{GauInit}  during growing; 
		(c) \textit{p-AutoGrow} with $K=50$; and (d) \textit{p-AutoGrow} with $K=3$.
		For better illustration, the dots on the trajectory are plotted every $4$, $20$, $5$ and $3$ epochs in (a-d), respectively.}
	\label{fig:trajectory-autogrow}
\end{figure*}
In this section, we study Network Morphism itself and its integration into our \autogrow{} under Convergent Growth.
When studying Network Morphism, we take the following steps:
1) train a shallower \resnet{} to converge,  
2) stack residual blocks on top of each sub-network to morph to a deeper net, 
3) use \texttt{ZeroInit} or \texttt{AdamInit} to initialize new layers, and
4) train the deeper net in a standard way. 
We compare the accuracy difference (``$\Delta$'') between Network Morphism and training the deeper net from scratch.
Table~\ref{tab:morphism-one2one} summaries our results.
Network Morphism has a lower accuracy (negative ``$\Delta$'') in all the cases, which validates our hypothesis that a converged shallow network with Network Morphism gives a bad initialization to train a deeper net.

We visualize the optimization trajectories to illustrate the hypothesis.
We hypothesize that a converged shallower net may not be an adequate initialization.
Figure~\ref{fig:trajectory-morphism} visualizes and compares the optimization trajectories of Network Morphism and the training from scratch.
In this figure, the shallower net is \texttt{Basic3ResNet-3-3-3} (\resnet{}-20) and the deeper one is \texttt{Basic3ResNet-5-5-5} (\resnet{}-32) in Table~\ref{tab:morphism-one2one}. 
The initializer is \texttt{ZeroInit}. 
The visualization method is extended from~\cite{li2018visualizing}.
Points on the trajectory are evenly sampled every a few epochs.
To maximize the variance of trajectory, we use PCA to project from a high dimensional space to a 2D space and use the first two Principle Components (PC) to form the axes in Figure~\ref{fig:trajectory-morphism}. 
The contours of training loss function and the trajectory are visualized around the final minimum of the deeper net.
When projecting a shallower net to a deeper net space, zeros are padded for the parameters not existing in the deeper net.
We must note that the loss increase along the trajectory does not truly represent the situation in high dimensional space, as the trajectory is just a projection.
It is possible that the loss remains decreasing in the high dimension while it appears in an opposite way in the 2D space. 
The sharp detour at ``Morphing'' in Figure~\ref{fig:trajectory-morphism}(a) may indicate that the shallower net plausibly converges to a point that the deeper net struggles to escape.
In contrast, Figure~\ref{fig:trajectory-morphism}(b) shows that the trajectory of the direct optimization in the deeper space smoothly converges to a better minimum.



To further validate our hypothesis, we integrate Network Morphism as the initializer in \autogrow{} with Convergent Growth policy. We refer to this version of \autogrow{} as \textit{c-AutoGrow} with ``\textit{c-}'' denoting ``Convergent.''
More specific, we take \texttt{ZeroInit} or \texttt{AdamInit} as sub-module initializer and ``Convergent Growth'' policy in Algorithm~\ref{algo:autogrow}.  
To recap, in this setting, \autogrow{} trains a shallower net till it converges, then grows a sub-module which is initialized by Network Morphism, and repeats the same process till there is no further accuracy improvement.
In every interval of $K$ training epochs (\texttt{train($g(\mathcal{X}_0), K$)} in Algorithm~\ref{algo:autogrow}), ``staircase'' learning rate is used. 
The learning rate is reset to $0.1$ at the first epoch, and decayed by $0.1 \times$ at epoch $\frac{K}{2}$ and $\frac{3K}{4}$.
The results are shown in Table~\ref{tab:autogrow-morphism} by ``\underline{\textit{staircase}}'' rows, 
which illustrate that 
\textit{c-AutoGrow} can grow a DNN multiple times and finally find a depth.
However, there are two problems:
1) the final accuracy is lower than training the found net from scratch, as indicated by ``$\Delta$'', validating our hypothesis; 
2) the depth learning stops too early with a relatively shallower net, while a deeper net beyond the found depth can achieve a higher accuracy as we will show in Table~\ref{tab:autogrow-adaptation}.
These problems provide a circumstantial evidence of the hypothesis that a converged shallow net with Network Morphism gives a bad initialization.
Thus, \autogrow{} cannot receive signals to continue growing after a limited number of growths.

Figure~\ref{fig:trajectory-autogrow}(a) visualizes the trajectory of \textit{c-AutoGrow} corresponding to row ``\texttt{2-3-6}'' in Table~\ref{tab:autogrow-morphism}. 
Along the trajectory, there are many trials to detour and escape an initialization from a shallower net.

\subsection{Ablation Study for \textbf{\autogrow{}} Design}
\label{exp:autogrow}

\begin{table}
  \caption{ \textit{p-AutoGrow} with different growing interval $K$. }
  \label{tab:autogrow-vs-gi}
  \centering
  \resizebox{.85\columnwidth}{!}{
  	\begin{tabular}{ccc}
  		\toprule
  		\multicolumn{3}{c}{CIFAR10} \\
  		$K$ & found net\textsuperscript{$\dagger$} &  accu \%  \\
  		\midrule
  		50 & \texttt{6-5-3}   & 92.95 \\
  		20 &  \texttt{7-7-7}  & 93.26 \\
  		10 &  \texttt{19-19-19}  & 93.46 \\
  		5  & \texttt{23-22-22}  & 93.98 \\
  		\textbf{3}  &  \textbf{\texttt{42-42-42}} & \textbf{94.27} \\
  		1  &  \texttt{77-76-76} & 94.30 \\
%
%
%
  		\bottomrule
  		\multicolumn{3}{l}{\textsuperscript{$\dagger$} \texttt{Basic3ResNet} }
  	\end{tabular}
  
  \quad
  
  \begin{tabular}{ccc}
  	\toprule
  	\multicolumn{3}{c}{CIFAR100} \\
  	$K$ & found net\textsuperscript{$\dagger$} &  accu \%  \\
  	\midrule 
  	50  & \texttt{8-5-7} & 72.07 \\
    20  & \texttt{8-11-10}  & 72.93 \\
  	10  &  \texttt{18-18-18} & 73.64 \\
  	5  &  \texttt{23-23-23} & 73.70 \\
  	\textbf{3}  &  \textbf{\texttt{54-53-53}} & \textbf{74.72} \\
  	1  &  \texttt{68-68-68} & 74.51 \\

  	\bottomrule
  	\multicolumn{3}{l}{\textsuperscript{$\dagger$} \texttt{Basic3ResNet} }
  \end{tabular}

  }
\end{table}

%
%
%

\begin{table}
  \caption{ \textit{p-AutoGrow} under initializers with $K=3$. }
  \label{tab:pautogrow-vs-inits}
  \centering
  \resizebox{1.\columnwidth}{!}{
  	\begin{tabular}{ccc}
  		\toprule
  		\multicolumn{3}{c}{CIFAR10} \\
  		initializer &  found net\textsuperscript{$\dagger$} & accu \\
  		\midrule 
  		\texttt{ZeroInit}  & \texttt{31-30-30} & 93.57   \\
  		\texttt{AdamInit} & \texttt{37-37-36} & 93.79  \\
  		\texttt{UniInit}  & \texttt{28-28-28} & 93.82  \\
  		\textbf{\texttt{GauInit}}  & \textbf{\texttt{42-42-42}} & \textbf{94.27}   \\

  		\bottomrule
		\multicolumn{3}{l}{\textsuperscript{$\dagger$} \texttt{Basic3ResNet} }
  	\end{tabular}
  
  \quad
  
  \begin{tabular}{ccc}
  	\toprule
  	\multicolumn{3}{c}{CIFAR100} \\
  	 initializer &  found net\textsuperscript{$\dagger$} & accu  \\
  	\midrule 
  	\texttt{ZeroInit}  & \texttt{26-25-25}  & 73.45 \\
  	\texttt{AdamInit}  & \texttt{27-27-27} & 73.92 \\
  	\texttt{UniInit}   & \texttt{41-41-41} & 74.31 \\
  	\textbf{\texttt{GauInit}}   & \textbf{\texttt{54-53-53}} & \textbf{74.72} \\

  	\bottomrule
  	\multicolumn{3}{l}{\textsuperscript{$\dagger$} \texttt{Basic3ResNet} }
  \end{tabular}
  }
\end{table} 

%
%
%

\begin{figure}
	\centering
	\includegraphics[width=1.0\columnwidth]{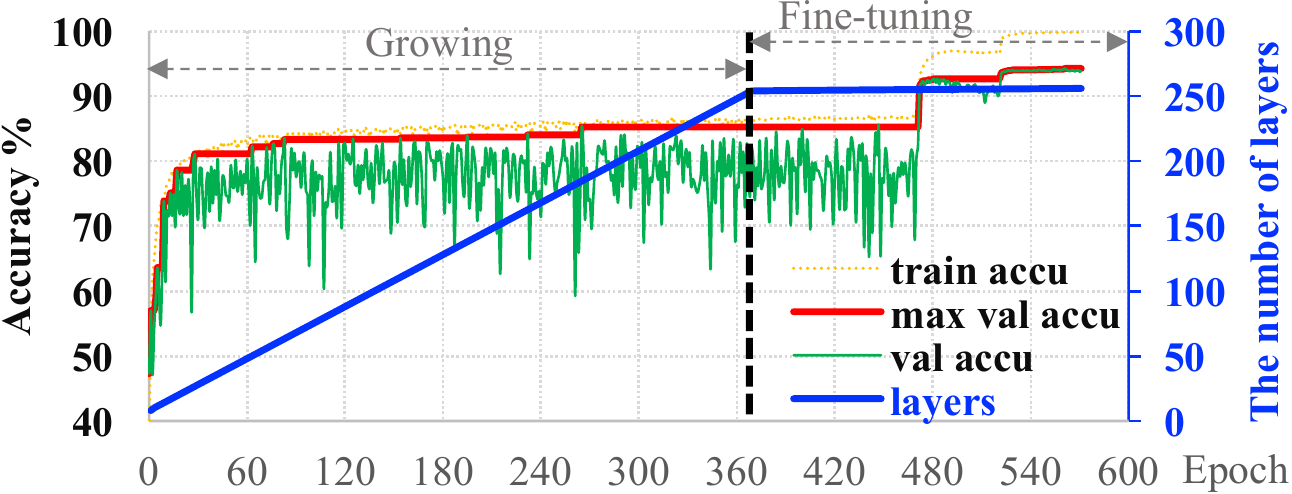}
	\captionof{figure}{\textit{p-AutoGrow} on CIFAR10 ($K=3$). The seed net is \texttt{Basic3ResNet-1-1-1}.}
	\label{fig:curve-cifar10-gi3}
\end{figure}

Based on the findings in Section~\ref{exp:morphism}, we propose three simple but effective solutions to further enhance \autogrow{} and refer it as \textit{p-AutoGrow}, with ``\textit{p-}'' denoting ``Periodic'':
%
%
\begin{itemize}
	\item Use a \textit{large constant} learning rate for growing, \textit{i.e.}, $0.1$ for residual networks and $0.01$ for plain networks. 
	Stochastic gradient descent with a large learning rate intrinsically introduces noises, which help to avoid a full convergence into a bad initialization from a shallower net. 
	Note that staircase learning rate is still used for fine-tuning after discovering the final DNN;
	\item Use \textit{random} initialization (\texttt{UniInit} or \texttt{GauInit}) as noises to escape from an inadequate initialization;
	\item Grow rapidly \textit{before} a shallower net converges by taking Periodic Growth with a small $K$.
\end{itemize}

\textit{p-AutoGrow} is our final \autogrow{}. In the rest part of this section, we perform ablation study to prove that the three solutions are effective.
%
We start from \textit{c-AutoGrow}, and incrementally add above solutions one by one and eventually obtain \textit{p-AutoGrow}.
In Table~\ref{tab:autogrow-morphism}, first, we replace the staircase learning rate with a constant large learning rate, the accuracy of \textit{AutoGrow} improves and therefore ``$\Delta$'' improves;
second, we further replace Network Morphism (\texttt{ZeroInit} or \texttt{AdamInit}) with a random initializer (\texttt{UniInit} or \texttt{GauInit}) and result in a bigger gain. 
Overall, combining a constant learning rate with \texttt{GauInit} performs the best.
Thus, constant learning rate and \texttt{GauInit} are adopted in the remaining experiments, unless we explicitly specify them. 
Figure~\ref{fig:trajectory-autogrow}(b) visualizes the trajectory corresponding to row ``\texttt{2-4-3}'' in Table~\ref{tab:autogrow-morphism}, which is much smoother compared to Figure~\ref{fig:trajectory-autogrow}(a), implying the advantages of constant large learning rate and \texttt{GauInit}.

Note that, in this paper, we are more interested in automating depth discovery to find a final DNN (``found net'') with a high accuracy (``accu''). Ideally, the ``found net'' has a minimum depth, a larger depth than which cannot further improve ``accu''. 
We will show in Figure~\ref{fig:autogrow-vs-from-scratch} that \autogrow{} discovers a depth approximately satisfying this property.
The ``$\Delta$'' is a metric to indicate how well shallower nets initialize deeper nets; a negative ``$\Delta$'' indicates that weight initialization from a shallower net hurts training of a deeper net; while a positive ``$\Delta$'' indicates \autogrow{} helps training a deeper net, which is a byproduct of this work.

Finally, we apply the last solution -- Periodic Growth, and obtain our final \textit{p-AutoGrow}.
Our ablation study results for \textit{p-AutoGrow} are summarized in Table~\ref{tab:autogrow-vs-gi}.
Table~\ref{tab:autogrow-vs-gi} analyzes the impact of the growing period $K$.
In general, $K$ is a hyper-parameter to trade off speed and accuracy:
a smaller $K$ takes a longer learning time but discovers a deeper net, vice versa.
Our results validate the preference of a faster growth (\ie{} a smaller $K$).
On CIFAR10/CIFAR100, the accuracy reaches plateau/peak at $K=3$;
further reducing $K$ produces a deeper net while the accuracy gain is marginal/impossible.
In the following, we simply select $K=3$ for robustness test. 
More importantly, our quantitative results in Table~\ref{tab:autogrow-vs-gi} show that \textit{p-AutoGrow} finds much deeper nets, overcoming the very-early stop issue in \textit{c-AutoGrow} in Table~\ref{tab:autogrow-morphism}. That is, Periodic Growth proposed in this work is much more effective than Convergent Growth utilized in previous work.


	Figure~\ref{fig:trajectory-autogrow}(c)(d) visualize the trajectories of \textit{p-AutoGrow} with $K=50$~and~$3$. 
	The 2D projection gives limited information to reveal the advantages of \textit{p-AutoGrow} comparing to \textit{c-AutoGrow} in Figure~\ref{fig:trajectory-autogrow}(b), although the trajectory of our final \textit{p-AutoGrow} in Figure~\ref{fig:trajectory-autogrow}(d) is plausibly more similar to the one of training from scratch in Figure~\ref{fig:trajectory-morphism}(b).

For sanity check, we perform the ablation study of initializers for \textit{p-AutoGrow}. 
The results are in Table~\ref{tab:pautogrow-vs-inits}, which further validates our wisdom on selecting \texttt{GauInit}.
The motivation of Network Morphism in previous work
was to start a deeper net from a loss function that has been well optimized by a shallower net, so as not to restart the deeper net 
training from scratch~\citep{wei2016network, wei2017modularized, chen2015net2net, elsken2017simple, cai2018efficient, cai2018path}. 
In all our experiments, we find that simple random initialization can achieve the same goal.
Figure~\ref{fig:curve-cifar10-gi3} plots the convergence curves and learning process for ``\texttt{42-42-42}'' in Table~\ref{tab:autogrow-vs-gi}. 
Even with \texttt{GauInit}, the loss and accuracy rapidly recover and no restart is observed.
The convergence pattern in the ``Growing'' stage is similar to the ``Fine-tuning'' stage under the same learning rate (the initial learning rate $0.1$). 
Similar results on ImageNet will be shown in Figure~\ref{fig:imagenet-curves}.
Our results challenge the necessity of Network Morphism when growing neural networks.

At last, we perform the ablation study on the initial depth of the seed network.
Table~\ref{tab:autogrow-vs-seed} demonstrates that a shallowest DNN works as well as a deeper seed.
This implies that \autogrow{} can appropriately stop regardless of the depth of the seed network.
As the focus of this work is on depth automation, we prefer starting with the shallowest seed to avoid a manual search of a seed depth.

\begin{table}
  \caption{ \textit{p-AutoGrow} with different seed architecture. }
  \label{tab:autogrow-vs-seed}
  \centering
  \resizebox{.85\columnwidth}{!}{
  	\begin{tabular}{cccc}
  		\toprule
  		dataset & seed net\textsuperscript{$\dagger$} & found net\textsuperscript{$\dagger$} &  accuracy \%  \\
  		\midrule 
  		\multirow{2}{*}{CIFAR10} & \texttt{1-1-1} & \texttt{42-42-42}   & 94.27 \\
  		& \texttt{5-5-5} & \texttt{46-46-46} & 94.16 \\
  		
  		\midrule 
  		\multirow{2}{*}{CIFAR10} & \texttt{1-1-1-1} & \texttt{22-22-22-22} & 95.49  \\
  		& \texttt{5-5-5-5} & \texttt{23-22-22-22} &  95.62 \\


  		\bottomrule
  		\multicolumn{4}{l}{\textsuperscript{$\dagger$} \texttt{Basic3ResNet} or \texttt{Basic4ResNet}. }
  	\end{tabular}
  }
\end{table}

\begin{table}
  \caption{ \autogrow{} improves accuracy of plain nets. }
  \label{tab:autogrow-plainnets}
  \centering
  \resizebox{.9\columnwidth}{!}{
  	\begin{tabular}{crccc}
  		\toprule
  		dataset & net & layer \# & method  & accu \%  \\
  		\midrule 
  		\multirow{4}{*}{CIFAR10} & \texttt{Plain4Net-6-6-6-6} & 26 & baseline  & 93.90 \\
  		\cmidrule{2-5}
  		  & \texttt{Plain4Net-6-6-6-6} & 26 & \begin{tabular}{@{}c@{}}\autogrow{} \\ $K=30$\end{tabular}  & 95.17 \\
  		  \cmidrule{2-5}
  		  & \texttt{Basic4ResNet-3-3-3-3} & 26 & baseline  & 95.33 \\
  		
  		\midrule
  		\multirow{4}{*}{CIFAR10} & \texttt{Plain3Net-11-11-10} & 34  & baseline & 90.45 \\
  		\cmidrule{2-5}
  		 & \texttt{Plain3Net-11-11-10} & 34 & \begin{tabular}{@{}c@{}}\autogrow{} \\ $K=50$\end{tabular} & 93.13 \\
  		 \cmidrule{2-5}
  		 & \texttt{Basic3ResNet-6-6-5} & 36 & baseline & 93.60 \\

  		\bottomrule
  		
  	\end{tabular}
  }
\end{table}

\subsection{Adaptability of \textbf{\autogrow{}}}
\label{exp:adaptation}

\begin{table*}
  \caption{ The adaptability of \autogrow{} to datasets}
  \label{tab:autogrow-adaptation}
  \centering
  \resizebox{2.\columnwidth}{!}{
  	\begin{tabular}{llllr}
  		\toprule
  		net & dataset &  found net & accu \%  & $\Delta$\textsuperscript{$*$} \\
  		\midrule 
  		\multirow{5}{*}{\texttt{Basic3ResNet}} & CIFAR10  & \texttt{42-42-42} & 94.27 & -0.03 \\
  		& CIFAR100  & \texttt{54-53-53} & 74.72 & \textbf{-0.95} \\
  		& SVHN  & \texttt{34-34-34} & 97.22 & 0.04 \\
  		& FashionMNIST  & \texttt{30-29-29} & 94.57 &  -0.06 \\
  		& MNIST  & \texttt{33-33-33} & 99.64 & -0.03 \\ 
  		
  		\midrule 
  		\multirow{5}{*}{\texttt{Basic4ResNet}} & CIFAR10  & \texttt{22-22-22-22} & 95.49 & -0.10 \\
  		& CIFAR100  & \texttt{17-51-16-16} & 79.47 & \textbf{1.22} \\
  		& SVHN  & \texttt{20-20-19-19} & 97.32 & -0.08 \\
  		& FashionMNIST  & \texttt{27-27-27-26} & 94.62 & \textbf{-0.17}\\
  		& MNIST  & \texttt{11-10-10-10} & 99.66 & 0.01 \\ 
  		
  		\bottomrule
  		\multicolumn{5}{l}{\textsuperscript{$*$} $\Delta = $ (accuracy of \autogrow{}) $-$ (accuracy of training from scratch) }
  	\end{tabular}
  \quad
  \begin{tabular}{llllr}
  	\toprule
  	net & dataset &  found net & accu \%  & $\Delta$\textsuperscript{$*$} \\
  	
  	\midrule 
  	\multirow{5}{*}{\texttt{Plain3Net}} & CIFAR10  & \texttt{23-22-22} & 90.82 & \textbf{6.49}\\
  	& CIFAR100  & \texttt{28-28-27} & 66.34 & \textbf{31.53} \\
  	& SVHN  & \texttt{36-35-35} & 96.79 & \textbf{77.20} \\
  	& FashionMNIST  & \texttt{17-17-17} & 94.49 & \textbf{0.56} \\
  	& MNIST  & \texttt{20-20-20} & 99.66 & \textbf{0.12} \\

  	\midrule 
  	\multirow{5}{*}{\texttt{Plain4Net}} & CIFAR10  & \texttt{17-17-17-17} & 94.20 & \textbf{5.72} \\
  	& CIFAR100  & \texttt{16-15-15-15} & 73.91 & \textbf{29.34} \\
  	& SVHN  & \texttt{12-12-12-11} & 97.08 & \textbf{0.32} \\
  	& FashionMNIST  & \texttt{13-13-13-13} & 94.47 & \textbf{0.72} \\
  	& MNIST  & \texttt{13-12-12-12} & 99.57 & 0.03 \\

  	\bottomrule
  	\multicolumn{5}{l}{}
  \end{tabular}
  
  }
\end{table*} 
\begin{table*}
	\caption{ The adaptability of \autogrow{} to dataset sizes}
	\label{tab:autogrow-subset-datasets}
	\centering
	\resizebox{1.2\columnwidth}{!}{
		\begin{tabular}{ccc}
			\toprule
			\multicolumn{3}{c}{\texttt{Basic3ResNet} on CIFAR10} \\
			dataset size & found net &  accu \%  \\
			\midrule 
			100\% &  \texttt{42-42-42}  &   94.27   \\
			75\% &  \texttt{32-31-31}  &   93.54   \\
			50\% &  \texttt{17-17-17}  &   91.34  \\
			25\% &  \texttt{21-12-7}  &    88.18  \\
			
			\bottomrule
			
			\toprule
			\multicolumn{3}{c}{\texttt{Basic4ResNet} on CIFAR100} \\
			dataset size & found net &  accu \%  \\
			\midrule 
			100\% &  \texttt{17-51-16-16}  &   79.47   \\
			75\% &  \texttt{17-17-16-16}  &   77.26   \\
			50\% &  \texttt{12-12-12-11}  &   72.91   \\
			25\% &  \texttt{6-6-6-6}  &   62.53   \\
			\bottomrule
			
		\end{tabular}
		\quad
		\begin{tabular}{ccc}
			\toprule
			\multicolumn{3}{c}{\texttt{Plain3Net} on MNIST} \\
			dataset size & found net &  accu \%  \\
			\midrule 
			100\% &  \texttt{20-20-20}  &   99.66  \\
			75\% &  \texttt{12-12-12}  &   99.54   \\
			50\% &  \texttt{12-11-11}  &   99.46   \\
			25\% &  \texttt{10-9-9}  &   99.33   \\
			\bottomrule
			
			\toprule
			\multicolumn{3}{c}{\texttt{Plain4Net} on SVHN} \\
			dataset size & found net &  accu \%  \\
			\midrule 
			100\% &  \texttt{12-12-12-11}  &   97.08  \\
			75\% &  \texttt{9-9-9-9}  &   96.71   \\
			50\% &  \texttt{8-8-8-8}  &   96.37   \\
			25\% &  \texttt{5-5-5-5}  &   95.68   \\
			\bottomrule
			
		\end{tabular}
		
	}
\end{table*} 


\begin{figure}
	\centering
	\includegraphics[width=1.\columnwidth]{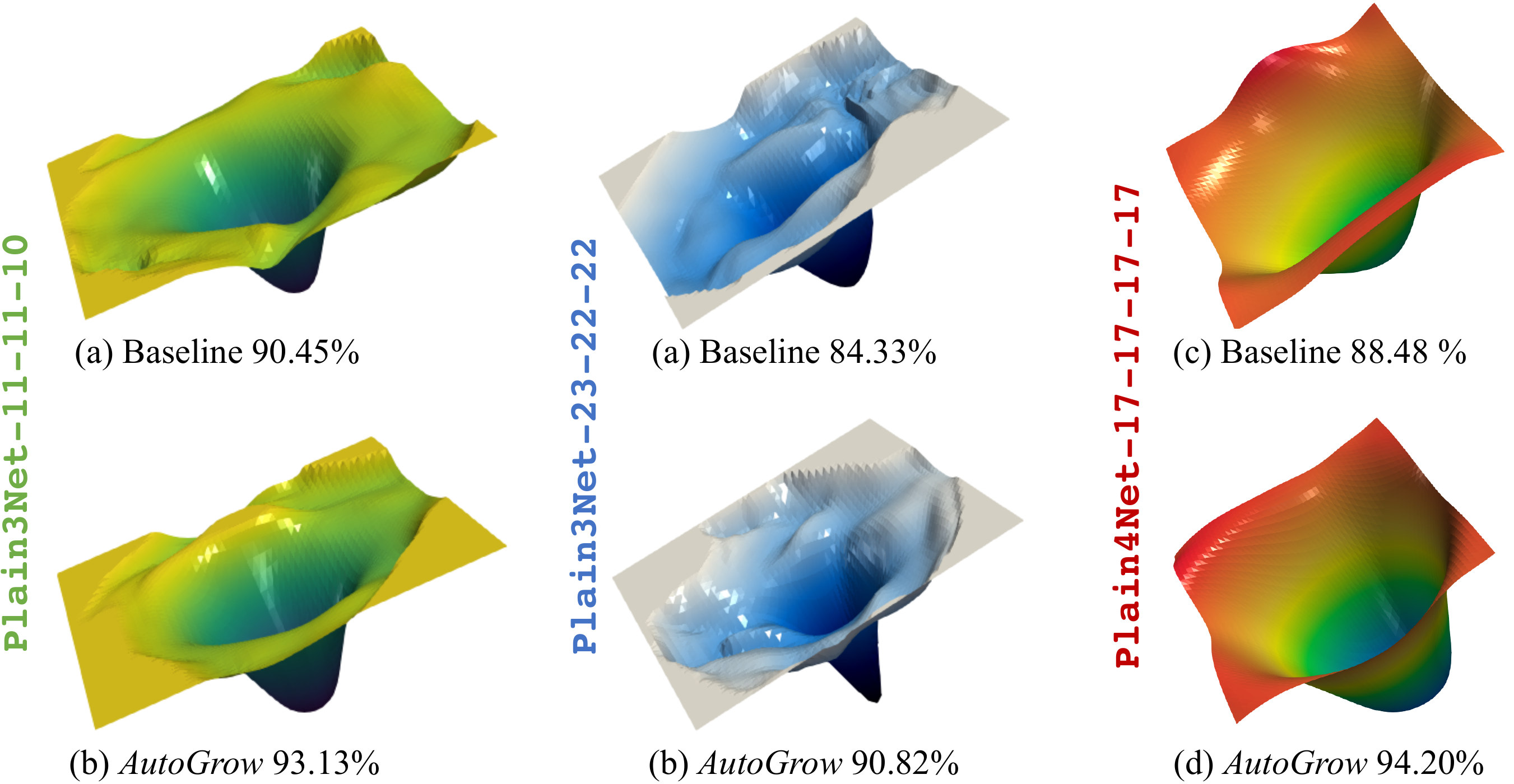}
	\caption{Loss surfaces around minima found by baselines and \autogrow{}. Dataset is CIFAR10.}
	\label{fig:autogrow-3d-contours}
\end{figure}

\begin{figure}
	\centering
	\small
	\includegraphics[width=1.0\columnwidth]{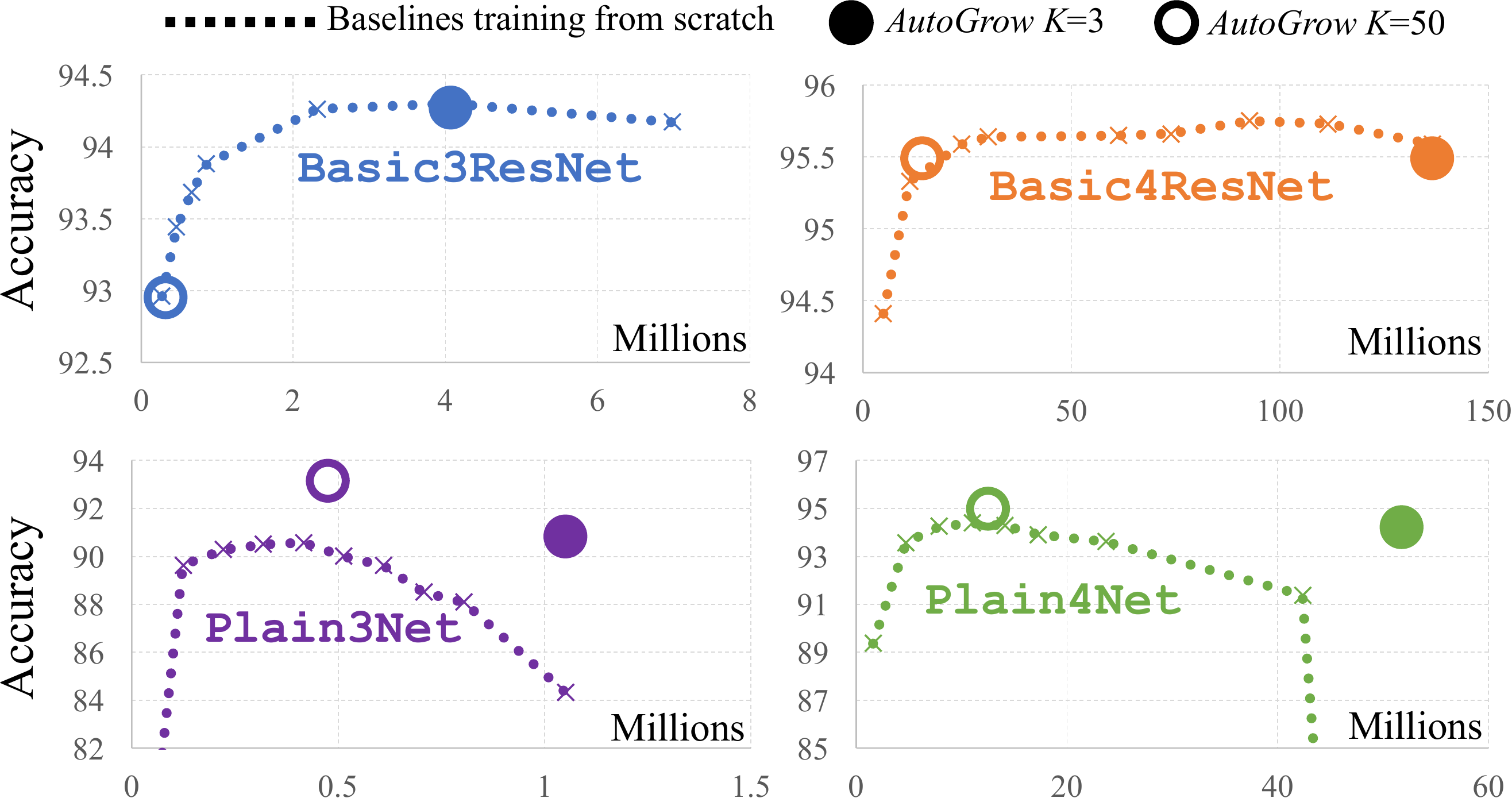}
	\captionof{figure}{\autogrow{} vs manual search obtained by training many baselines from scratch. $x-axis$ is the number of parameters. Dataset is CIFAR10.}
	\label{fig:autogrow-vs-from-scratch}
\end{figure}

To verify the adaptability of \autogrow{}, we use an identical configuration (\textit{p-AutoGrow} with $K=3$) and test over $5$ datasets and $4$ seed architectures. 
Table~\ref{tab:autogrow-adaptation} includes the results of all 20 combinations.
Figure~\ref{fig:autogrow-vs-from-scratch} compares \autogrow{} with manual search which is obtained by training many DNNs with different depths from scratch.
The results lead to the following conclusions and contributions:
\begin{enumerate}
\item In Table~\ref{tab:autogrow-adaptation}, \autogrow{} discovers layer depth across all scenarios without any tuning, achieving the main goal of this work.
Manual design needs $m \cdot n \cdot k$ trials, where $m$ and $n$ are respectively the numbers of datasets and sub-module categories, and $k$ is the number of trials per dataset per sub-module category;
	
\item For \resnets{}, a discovered depth (``$\CIRCLE$'' in Figure~\ref{fig:autogrow-vs-from-scratch}) falls at the location where accuracy saturates. 
This means \autogrow{} discovers a near-optimal depth: a shallower depth will lose accuracy while a deeper one gains little.
The final accuracy of \autogrow{} is as good as training the discovered net from scratch as indicated by ``$\Delta$'' in Table~\ref{tab:autogrow-adaptation}, indicating that initialization from shallower nets does not hurt training of deeper nets. 
As a byproduct, in plain networks, there are large positive ``$\Delta$''s in Table~\ref{tab:autogrow-adaptation}.
It implies that baselines fail to train very deep plain networks even using Batch Normalization, but \autogrow{} enables the training of these networks;
Table~\ref{tab:autogrow-plainnets} shows the accuracy improvement of plain networks by tuning $K$, approaching the accuracy of \resnets{} with the same depth.
Figure~\ref{fig:autogrow-3d-contours} visualizes loss surfaces around minima by \autogrow{} and baseline. 
Intuitively, \autogrow{} finds wider or deeper minima with less chaotic landscapes.

\begin{table*}
  \caption{ Scaling up to ImageNet}
  \label{tab:autogrow-imagenet}
  \centering
  \resizebox{1.2\columnwidth}{!}{
  	\begin{tabular}{llllcc}
  		\toprule
  		net & $K$ &  found net & Top-1  & Top-5 & \textsuperscript{$\dagger$}$\Delta$ Top-1  \\
  		\midrule 
  		\multirow{2}{*}{\texttt{Basic4ResNet}} & 2 & \texttt{12-12-11-11} & 76.28 & 92.79 & 0.43 \\
  		 & 5  & \texttt{9-3-6-4} & 74.75 & 91.97 & 0.72 \\
  		 
  		 \midrule 
  		 \multirow{2}{*}{\texttt{Bottleneck4ResNet}} & 2 & \texttt{6-6-6-17} & 77.99  & 93.91 & $0.83$\\
  		 & 5  & \texttt{6-7-3-9} & 77.33 & 93.65 & 0.83 \\
  		 
  		 \midrule 
  		 \multirow{2}{*}{\texttt{Plain4Net}} & 2 & \texttt{6-6-6-6} & 71.22  & 90.08 & 0.70\\
  		 & 5  & \texttt{5-5-5-4} & 70.54 & 89.76 & 0.93 \\

  		\bottomrule
  		\multicolumn{5}{l}{\textsuperscript{$\dagger$} $\Delta = $ (Top-1 of \autogrow{}) $-$ (Top-1 of training from scratch) }
  	\end{tabular}
  }
\end{table*} 
\begin{table*}
  \caption{The efficiency of \autogrow{}}
  \label{tab:autogrow-imagenet-timing}
  \centering
  \resizebox{1.2\columnwidth}{!}{
  	\begin{tabular}{rlrr}
  		\toprule
  		net & GPUs &  growing & fine-tuning \\
  		\midrule 
  		\multirow{1}{*}{\texttt{Basic4ResNet-12-12-11-11}} & $4$ GTX 1080 Ti & $56.7$ hours & $157.9$ hours\\
  		\texttt{Basic4ResNet-9-3-6-4} & $4$ GTX 1080 & $47.9$ hours & $65.8$ hours \\
  		 
  		 \midrule 
  		 \multirow{1}{*}{\texttt{Bottleneck4ResNet-6-6-6-17}} & $4$ TITAN V & $45.3$ hours & $114.0$ hours \\
  		 \texttt{Bottleneck4ResNet-6-7-3-9} & $4$ TITAN V & $61.6$ hours & $78.6$ hours\\
  		 
  		 \midrule 
  		 \multirow{1}{*}{\texttt{Plain4Net-6-6-6-6}} & $4$ GTX 1080 Ti & $11.7$ hours & $29.7$ hours\\
  		 \texttt{Plain4Net-5-5-5-4} & $4$ GTX 1080 Ti & $25.6$ hours & $25.3$ hours \\
  		 
  		\bottomrule
  	\end{tabular}
  }
\end{table*} 
	
\item For robustness and generalization study purpose, we stick to $K=3$ in our experiments, however, we can tune $K$ to trade off accuracy and model size. 
As shown in Figure~\ref{fig:autogrow-vs-from-scratch}, \autogrow{} discovers smaller DNNs when increasing $K$ from $3$ (``$\bm{\CIRCLE}$'') to $50$ (``$\bm{\Circle}$''). 
Interestingly, the accuracy of plain networks even increases at $K=50$. 
This implies the possibility of discovering a better accuracy-depth trade-off by tuning $K$,
although we stick to $K=3$ for generalizability study and it generalizes well.

\item In Table~\ref{tab:autogrow-adaptation}, \autogrow{} discovers different depths under different sub-modules. The final accuracy is limited by the sub-module design, not by our \autogrow{}. Given a sub-module architecture, our \autogrow{} can always find a near-optimal depth. 
\end{enumerate}

Finally, our supposition is that: when the size of dataset is smaller, the optimal depth should be smaller.
Under this supposition, we test the effectiveness of \autogrow{} by sampling a subset of dataset and verify if \autogrow{} can discover a shallower depth.
Table~\ref{tab:autogrow-subset-datasets} summarizes the results.
In each set of experiments, dataset is randomly down-sampled to $100\%$, $75\%$, $50\%$ and $25\%$.
For a fair comparison, $K$ is divided by the percentage of dataset such that the number of mini-batches between growths remains the same.
As expected, our experiments show that \autogrow{} adapts to shallower networks when the datasets are smaller. 


\subsection{Scaling to ImageNet and Efficiency}
\label{exp:imagenet}

\begin{figure}
	\centering
	\small
	\includegraphics[width=.9\columnwidth]{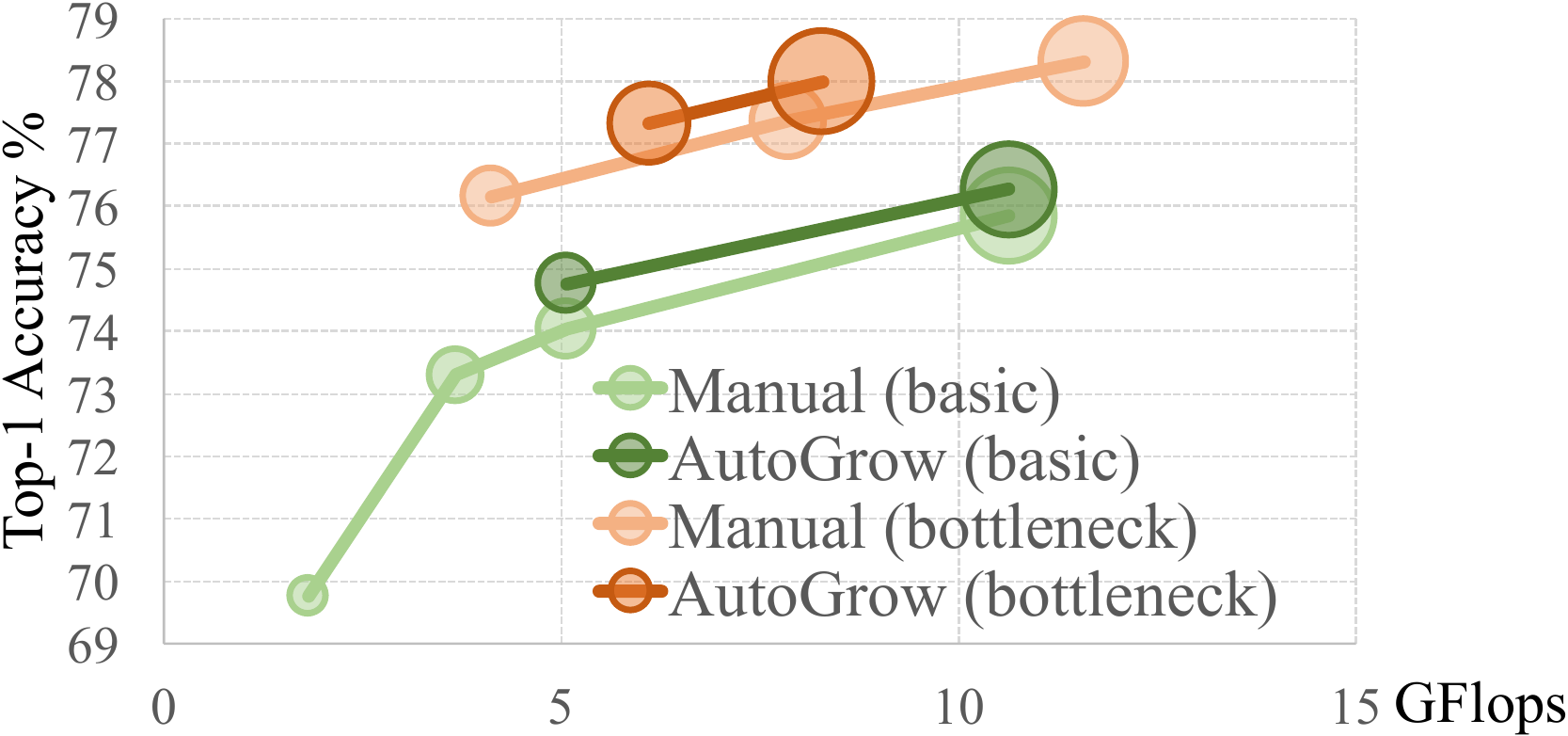}
	\captionof{figure}{\autogrow{} \textit{vs.} manual design~\citep{he2016deep} on ImageNet. 
		Marker area is proportional to model size determined by depth. 
		``basic''(``bottleneck'') refers to \resnets{} with basic (bottleneck) residual blocks.}
	\label{fig:imagenet}
\end{figure}

In ImageNet, $K=3$ should generalize well, but we explore \autogrow{} with $K=2$ and $K=5$ to obtain an accuracy-depth trade-off line for comparison with human experts.
The larger $K=5$ enables \autogrow{} to obtain a smaller DNN to trade-off accuracy and model size (computation) and the smaller $K=2$ achieves higher accuracy.
The results are shown in Table~\ref{tab:autogrow-imagenet}, which proves that \autogrow{} automatically finds a good depth without any tuning.
As a byproduct, the accuracy is even higher than training the found net from scratch, indicating that the Periodic Growth in \autogrow{} helps training deeper nets.
The comparison of \autogrow{} and manual depth design~\citep{he2016deep} is in Figure~\ref{fig:imagenet}, which shows that \autogrow{} achieves better trade-off between accuracy and computation (measured by floating point operations).

\begin{figure}
	\centering
	\includegraphics[width=1.0\columnwidth]{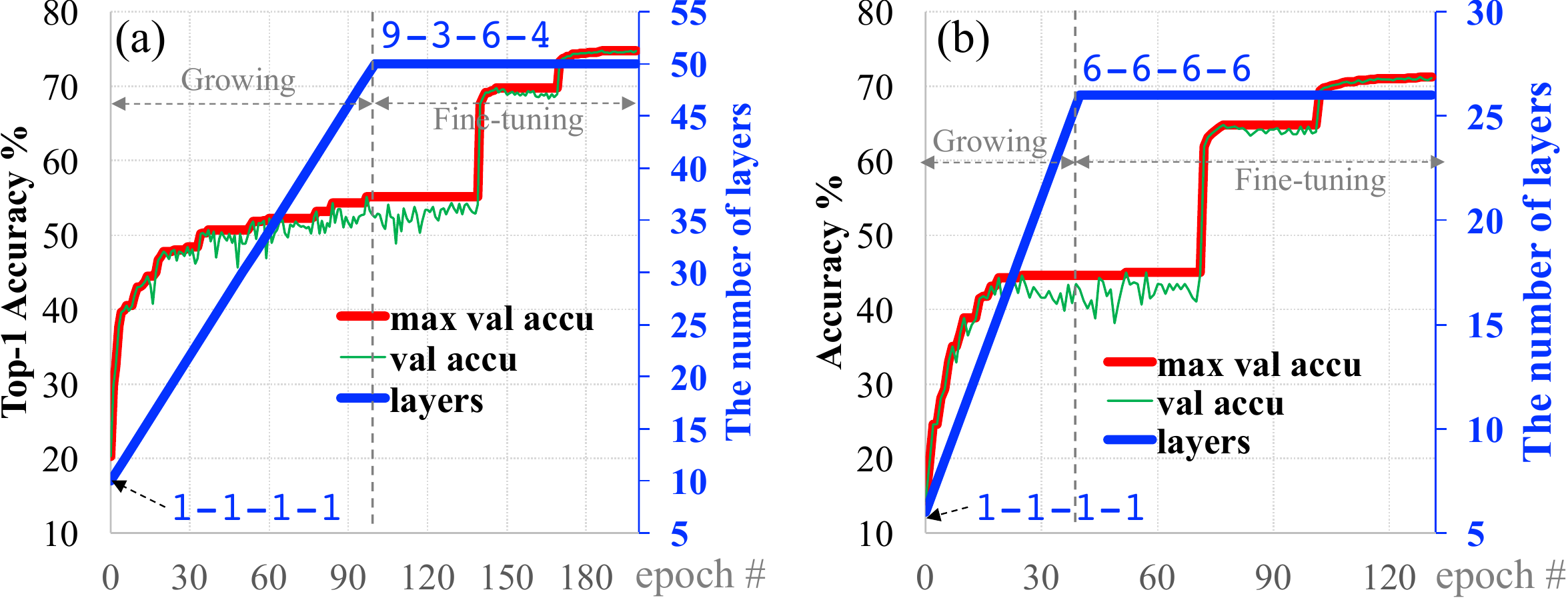}
	\vspace{-12pt}
	\caption{The convergence curves and growing process on ImageNet for (a) \texttt{Basic4ResNet-9-3-6-4} and (b) \texttt{Plain4Net-6-6-6-6} in Table~\ref{tab:autogrow-imagenet-timing}.}
	\label{fig:imagenet-curves}
\end{figure}

Table~\ref{tab:autogrow-imagenet-timing} summarizes the breakdown of wall-clock time in \autogrow{}. 
The growing/searching time is as efficient as (often more efficient than) fine-tuning the single discovered DNN. 
The scalability of \autogrow{} comes from its intrinsic features that (1) it grows quickly with a short period $K$ and stops immediately if no improvement is sensed; and (2) the network is small at the beginning of growing.
Figure~\ref{fig:imagenet-curves} plots the growing and converging curves for two DNNs in Table~\ref{tab:autogrow-imagenet-timing}. Even with random initialization in new layers, the accuracy converges stably.

\section{Related Work}
\label{sec:related}

Neural Architecture Search (NAS)~\citep{zoph2016neural} and neural evolution~\citep{miikkulainen2019evolving,angeline1994evolutionary,stanley2002evolving,liu2017hierarchical,real2017large} can search network architectures from a gigantic search space.
In NAS, the depth of DNNs in the search space is fixed, while \autogrow{} learns the depth.
Some NAS methods~\citep{bender2018understanding,liu2018darts, cortes2017adanet} can find DNNs with different depths, however, the maximum depth is pre-defined and shallower depths are obtained by padding zero operations or selecting shallower branches, while our \autogrow{} learns the depth in an open domain to find a minimum depth, beyond which no accuracy improvement can be obtained.
Moreover, NAS is very computation and memory intensive.
To accelerate NAS, one-shot models~\citep{saxena2016convolutional,pham2018efficient,bender2018understanding}, DARTS~\citep{liu2018darts} and NAS with Transferable Cell~\citep{zoph2018learning,liu2018progressive} were proposed.
The search time reduces dramatically but is still long from practical perspective.
It is still very challenging to deploy these methods to larger datasets such as ImageNet.
In contrast, our \autogrow{} can scale up to ImageNet thanks to its short depth learning time, which is as efficient as training a single DNN. 


In addition to architecture search which requires to train lots of DNNs from scratch, there are also many studies on learning neural structures within a single training. 
Structure pruning and growing were proposed for different goals, such as efficient inference~\citep{wen2016learning, park2016faster, li2016pruning, wen2017learning, yang2019deephoyer, lebedev2016fast, he2017channel, luo2017thinet, liu2017learning, dai2017nest, huang2018condensenet, gordon2018morphnet, du2019cgap}, lifelong learning~\citep{yoon2017lifelong} and model adaptation~\citep{feng2015learning, philipp2017nonparametric}.
However, those works fixed the network depth and limited structure learning within the existing layers.
Optimization over a DNN with fixed depth is easier as the skeleton architecture is known. 
\autogrow{} performs in a scenario where the DNN depth is unknown hence we must seek for the optimal depth.


\section{Conclusion}
\label{sec:conclusion}
In this paper, we propose a simple but effective layer growing algorithm (\autogrow{}) to automate the depth design of deep neural networks.
We empirically show that \autogrow{} can adapt to different datasets for different layer architectures without tuning hyper-parameters.
\autogrow{} can significantly reduce human effort on searching layer depth.
We surprisingly find that a rapid growing (under a large constant learning rate with random initialization of new layers) outperforms more intuitively-correct growing method, such as Network Morphism growing after a shallower net converged.
We believe our initiative results can inspire future research on structure growth of neural networks and related theory.

\subsubsection*{Acknowledgments}
This work was supported in part by NSF CCF-1725456, NSF 1937435, NSF 1822085, NSF CCF-1756013, NSF IIS-1838024 and DOE DE-SC0018064.
Any opinions, findings, conclusions or recommendations expressed
in this material are those of the authors and do not necessarily reflect the views of NSF, DOE or its contractors.
We also thank Dr. Yandan Wang (yandanw@unr.edu) at University of Nevada Reno for valuable feedback on this work.

\bibliographystyle{ACM-Reference-Format}
\bibliography{egbib}


\end{document}